\documentclass[conference]{IEEEtran}
\usepackage{cite}
\usepackage{amsmath,amssymb,amsfonts}
\usepackage{algorithmic}
\usepackage{graphicx}
\usepackage{textcomp}
\usepackage{xcolor}
\usepackage{hyperref}
\usepackage{threeparttable}
\usepackage[inkscapelatex=false]{svg}
\usepackage{verbatim}
\hypersetup{
hidelinks,
colorlinks=true,
allcolors=blue,
}
\makeatletter
\newcommand{\linebreakand}{%
  \end{@IEEEauthorhalign}
  \hfill\mbox{}\par
  \mbox{}\hfill\begin{@IEEEauthorhalign}
}
\makeatother

\def\BibTeX{{\rm B\kern-.05em{\sc i\kern-.025em b}\kern-.08em
    T\kern-.1667em\lower.7ex\hbox{E}\kern-.125emX}}
\begin{document}

\title{SSformer: A Lightweight Transformer for Semantic Segmentation\\
}

\author{\IEEEauthorblockN{ Wentao Shi\textsuperscript{\textsection}}
\IEEEauthorblockA{\textit{College of Computer Science and Technology} \\
\textit{Nanjing University of Aeronautics and Astronautics}\\
Nanjing, China \\
shiwentao@nuaa.edu.cn}
%\and
\and
\IEEEauthorblockN{ Jing Xu\textsuperscript{\textsection}}
\IEEEauthorblockA{\textit{College of Computer Science and Technology} \\
\textit{Nanjing University of Aeronautics and Astronautics}\\
Nanjing, China \\
jing.xu@nuaa.edu.cn
}
\linebreakand

\IEEEauthorblockN{Pan Gao}
\IEEEauthorblockA{\textit{College of Computer Science and Technology} \\
\textit{Nanjing University of Aeronautics and Astronautics}\\
pan.gao@nuaa.edu.cn}
}

\maketitle
\begingroup\renewcommand\thefootnote{\textsection}
\footnotetext{Equal contribution}
\endgroup

\begin{abstract}
It is well believed that Transformer performs better in semantic segmentation compared to convolutional neural networks. Nevertheless, the original Vision Transformer \cite{b2} may lack of inductive biases of local neighborhoods and possess a high time complexity. Recently, Swin Transformer \cite{b3} sets a new record in various vision tasks by using hierarchical architecture and shifted windows while being more efficient. However, as Swin Transformer is specifically designed for image classification, it may achieve suboptimal performance on dense prediction-based segmentation task. Further, simply combing Swin Transformer with existing methods would lead to the boost of model size and parameters for the final segmentation model. In this paper, we rethink the Swin Transformer for semantic segmentation, and design a lightweight yet effective transformer model, called SSformer. In this model, considering the inherent hierarchical design of Swin Transformer, we propose a decoder to aggregate information from different layers, thus obtaining both local and global attentions. Experimental results show the proposed SSformer yields comparable mIoU performance with state-of-the-art models, while maintaining a smaller model size and lower compute. Source code and pretrained models are available at: \href{https://github.com/shiwt03/SSformer}{https://github.com/shiwt03/SSformer}
\end{abstract}


\section{Introduction}
The field of computer vision and multimedia has been largely changed since Transformer was applied in image or video from a sequence-to-sequence perspective\cite{b11}\cite{b12}. It challenged the dominance of convolution neural network in this field \cite{b6}. However, simply applying Transformer into semantic segmentation brings huge time and space complexity, which limits its wider appliance. 

To reduce computational complexity, in Swin Transformer, the number of tokens is reduced by patch merging layers as the network gets deeper, and the self-attention calculation is limited to the non-overlapping local window. Generally speaking, Swin Transformer is linearly growing with the size of the image while the compute complexity of the original Vision Transformer (ViT) is of quadratic growth \cite{b2}. 

Swin Transformer is able to serve as a general-purpose backbone for computer vision. It has already been applied in detection, classification and segmentation and obtains strong performances. When Swin Transformer is applied to semantic segmentation, it is usually equipped with Upernet as its decoder, which can bring the mIoU value as high as 53.5 on ADE20K dataset. However, simply combining the Swin Transformer with an existing segmentation head would result in huge compute complexity increase and model parameter increase, since the segmentation head in existing methods is generally sophisticated and has not been optimized for transformer. 

Inspired by the fact that hierarchical design and the shifted window approach also prove beneficial for all-MLP architectures, in this paper, we design a  lightweight multilayer perceptron (MLP) decoder specially for Swin Transformer for image segmentation, in expectation of getting a model that balances performance and efficiency. More specifically, based on the intrinsic characteristics of Swin Transformer of providing multi-scale features, we proposed a decoder to fuse the features output from the different stages. The fused features contain high-resolution coarse feature and low-resolution fine feature, thus providing powerful representation for sematic segmentation. Note that, as we only introduce a simple MLP decoder to Transformer, it brings little complexity and parameter increase. Importantly, our proposed SSformer makes fine-tuning for semantic segmentation compatible with Swin Transformer pretraining.

The rest of the paper is organized as follows. Section \ref{Reated_work} reviews some recent deep learning based semantic segmentation models. Section \ref{Proposed_model} presents our proposed lightweight segmentation model. Experimental results and analysis are provided in Section \ref{experiment}, followed by concluding remarks in Section \ref{conclusion}.

\section{Related work}
\label{Reated_work}

\begin{figure*}[ht]
\includegraphics[width=\textwidth]{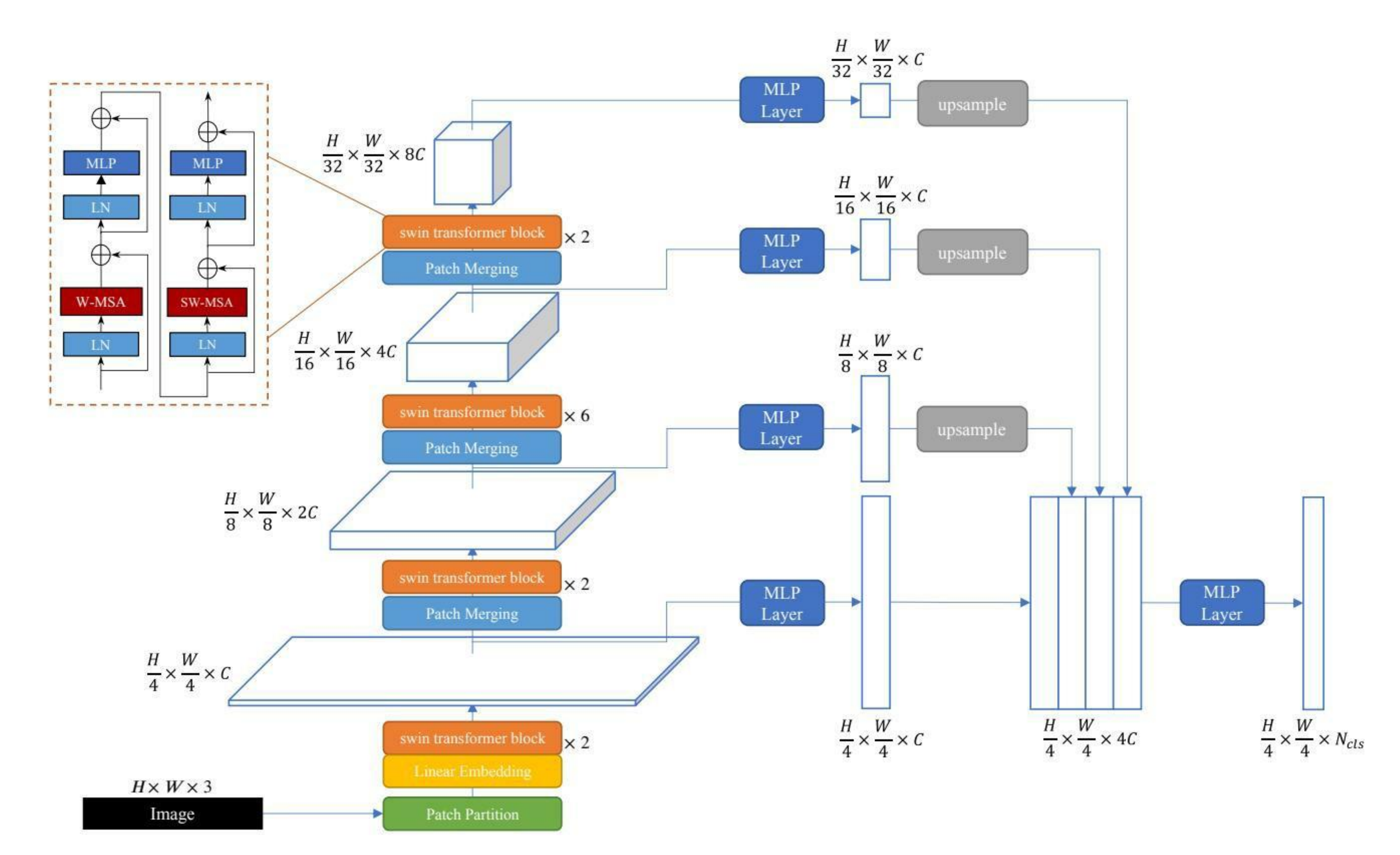}
\caption{The proposed SSformer architecture consists of two modules. A Swin Transformer is used to extract multi-level features; and a lightweight All-MLP decoder is designed to directly fuse these hierarchical features, which are finally upsampled to get the output. (It is omitted in this figure for clarity)}
\label{fig1}
\end{figure*}

\subsection{Fully Convolutional Networks (FCN) }

FCN is a classic method of semantic segmentation\cite{b8}. This work is considered a milestone in image segmentation, demonstrating that deep networks can be trained for semantic segmentation in an end-to-end manner on variable sized images \cite{b6}.

\subsection{Transformer }

In recent years, as an effectual method in natural language processing (NLP) field, Transformer was successfully transferred to computer vision (CV) domain, including image classification, semantic segmentation, etc\cite{b2}. Transformer is able to model the long-range dependency due to the self-attention mechanism. Vision Transformer (ViT) has convincingly proved that a pure transformer backbone can deal with all kinds of missions in CV. But ViT requires onerous training and possesses high compute complexity, which limited its wider appliance \cite{b2}.

\subsection{Swin Transformer }
The emerging of Swin Transformer effectively fills the gap between Transformer and semantic segmentation, by the reason that Swin Transformer adds an inductive bias to make Transformer better adapt to dense prediction tasks of semantic segmentation. It disperses more attention to the connection of each patch by the application of shifted window and patch merging\cite{b3}. 

\subsection{SETR }
SETR deploys an original transformer to encode images as a sequence of patches. ViT is the first work showing that a pure Transformer based image classification model can yield the state-of-the-art performance, while SETR is the first work to show that a pure Transformer can also be the-state-of the-art structure in image segmentation\cite{b1}. SETR further proves that Transformer can make advancement in the field of semantic segmentation. However, due to the use of original Transformer and redundant decoders, SETR has huge compute complexity and vast model parameters, which are often unacceptable in practice. As a result, the main purpose of our work is to reduce compute complexity and model parameters.

\section{The Proposed SSformer}
\label{Proposed_model}
In this section we will introduce SSformer, our simplified but still powerful semantic segmentation architecture. As illustrated in Figure \ref{fig1}, The proposed SSformer architecture consists of two modules. A Swin Transformer is employed to extract multi-level features; and a lightweight All-MLP decoder is proposed to fuse these hierarchical features. The fused features are finally upsampled to get the output.

In our implementation, we follow the original version of Swin Transformer which does a series of steps to shrink the size of image to be utilized by the Swin Transformer block. A linear embedding layer is applied on the shallow feature to project it to an arbitrary dimension (denoted as $C$). The Swin Transformer block has been concretely illustrated in the paper of Swin Transformer, so we don’t repeat it here. After the Swin Transformer block, a patch merging is used. It performs like pooling to down-sample the feature image.

Supposing that an image of size $H*W*C$ is given, $H$ and $W$ are the height and width of the image, respectively. After a series of processing, the size before patch merging is $\frac{H}{4}*\frac{W}{4}*C$. In the first patch merging, it uses a function like Dilated Convolution (but without any parameter) to exchange for more channels with the dimension of space. Then the tensor is shrunk to $\frac{H}{8}*\frac{W}{8}*2C$. In order to keep the number of channels being the same as convolutional neural networks, it uses a 1×1 convolution to lift the channels to $2C$. As a result, we finally change a tensor of $\frac{H}{4}*\frac{W}{4}*C$ into a tensor of $\frac{H}{8}*\frac{W}{8}*2C$. The next three patch merging operations are similar, so we just take the first one as an example.

SSformer incorporates an all-MLP decoder, which is made up only of the MLP layers. This kind of decoder was firstly used in Segformer architecture \cite{b4}, whose backbone is a Mix Transformer and it has a large effective receptive field (ERF). Nevertheless, the Mix Transformer is a redesigned hierarchical pretraining structure, which incorporates additional convolutional layers and needs to be trained from scratch.  In contrast, in SSformer, we use the original Swin Transformer as backbone, which has larger effective receptive field than CNN encoders or even tie with pure Transformer, while possessing smaller size of model due to its characteristics of hierarchical architecture by merging image patches and using shifted window. At the decoder, we use an MLP layer to map the output patch embedding of each stage to a feature map of channel dimension of $C$ . Then, the low-resolution feature maps are upsampled to the size as the feature map of the first stage and concatenated together. Finally, we use an MLP layer to transform the fused feature maps to segmentation masks of dimension of $ \frac{H}{4}*\frac{W}{4}*N_{cls} $, where $ N_{cls} $ represents the number of categories.  As a result, we believe that such a simple decoder can match the Swin Transformer based encoder well. Our proposed SSformer enables the original Swin Transformer architecture to be fine-tuned for semantic segmentation without the need to redesign a hierarchical backbone for pre-training.

Based on the proposed  architecture presented in Fig. 1, the computational complexity of SSformer on an image of $h \times w$ patches is as follows
\begin{equation}
\Omega(SSformer)=4hwC^2+2M^2hwC+hwC^2+4hwCN_{cls}
\label{complexity_eqn}
\end{equation}
where $4hwC^2+2M^2hwC$ represnets the complexity of the Swin Transformer based encoder, and $hwC^2+4hwCN_{cls}$ is the complexity induced by our proposed MLP decoder. In the above equation, $hw$ denotes the total number of patches in an image, where if the patch size is $P \times P$, then, $h=\frac{H}{P}$ and $w=\frac{W}{P}$. $M$ represents the window size used in the Swin Transformer. As can be observed from \eqref{complexity_eqn}, the computational complexity of our proposed SSformer is also linear with the input image size, which is thus a lightweight model in general. 

\section{Experiments}\label{experiment}
To demonstrate the effectiveness of the proposed SSformer model, we trained and tested SSformer with a single Nvidia RTX 3090Ti GPU. We use MMSegmentation as our code base and some data of other models is from MMSegmentation’s model zoo \cite{b7}.  More details about how to reproduce the results of the proposed SSformer can be found at \href{https://github.com/shiwt03/SSformer}{https://github.com/shiwt03/SSformer}. In what follows,  we compare SSformer with some other models on ADE20K and Cityscapes datasets.

\subsection{Experiments on ADE20K}
ADE20K is a widely used semantic segmentation dataset, covering a broad range of 150 semantic categories. Compared to other datasets, scene parsing for ADE20K is challenging due to the huge semantic concepts.  It has 25K images in total, with 20K for training, 2K for validation, and another 3K for testing. 
During implementation, we use AdamW as our optimizer. The channel numbers (i.e., the “C” dimension in Figure 1) after Linear Embedding was set to 128, which balances the performance and parameter amounts. The models are pretrained on ImageNet-22K, and fine-tuned with a batch size of 8 and the input images of 512×512 resolution for 160K iterations. The learning rate is set to 6e-5 at the beginning.
\begin{figure}[htbp]
\centering
\centerline{\includegraphics[width=\columnwidth]{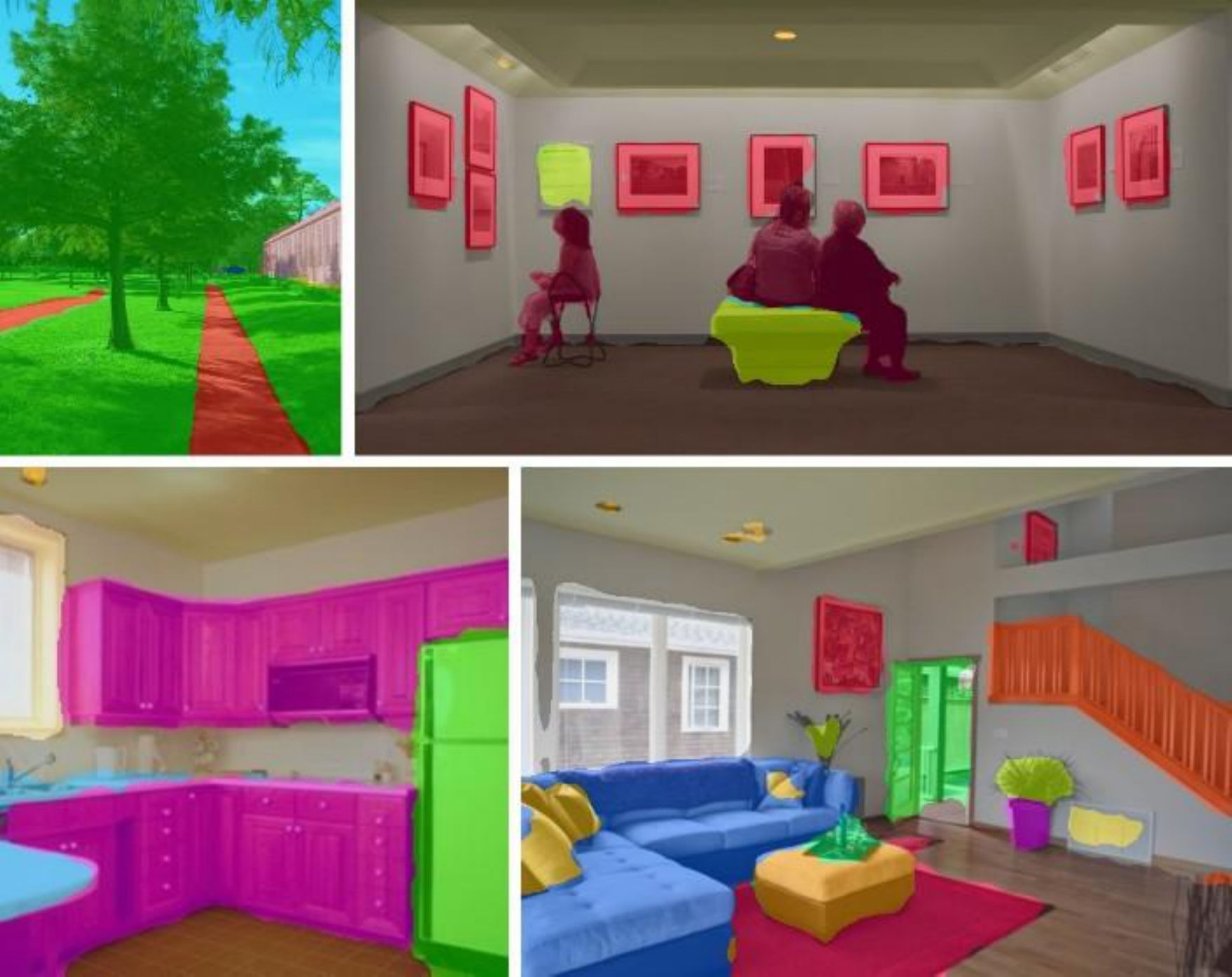}}
\caption{Part of segmentation results of SSformer on ADE20K}
\label{fig2}
\end{figure}

\begin{table}[htb]   
\begin{center}   
\caption{Comparison of segmentation results of different methods on ADE20K dataset. }
\label{table1} 
\begin{tabular}{l l l l l}   
\hline   Method & Iteration & Params & Flops* & mIoU \\  
\hline 
FCN\cite{b8} & 160000 & 68.6M	& 275.7G & 39.91\\
DeepLabV3+\cite{b13} & 160000 & 62.7M & 255.1G & 45.47\\
DMnet\cite{b15} & 160000 & 72.3M & 273.6G & 45.42\\
PSPNet\cite{b9} & 160000 & 68.1M & 256.4G & 44.39\\
PSANet\cite{b14} & 160000 & 73.1M & 272.5G & 43.74\\
SETR-PUP\cite{b1} & 160000 & 317.3M & 362.1G & 48.24\\
Swin-T\cite{b3} & 160000 & 121.3M & 297.2G & 50.31\\
SSformer & 160000 & 87.5M & 91.01G & 47.71\\
\hline   
\end{tabular}   
\begin{tablenotes}
*The Flops are calculated with the input images of 512×512 resolution.
\end{tablenotes}
\end{center}   
\end{table}
As shown in Table \ref{table1}, compared with other mainstream works in semantic segmentation, SSformer has a considerably low time complexity and keeps a competitive performance at the same time.

\subsection{Experiments on Cityscapes}
Cityscapes is a large-scale dataset which mainly contains images of urban roads, streets and landscapes. It has 5K images in total, with 2975 for training, 500 for validation, and another 1525 for testing. 
We employ the same optimizer and channels as the ones we use in ADE20K. The models are pretrained on ImageNet-22K, and fine-tuned with a batch size of 2 and the input images of 1024x1024 resolution for 80K iterations. 

\begin{figure}[htbp]
\centering
\centerline{\includegraphics[width=\columnwidth]{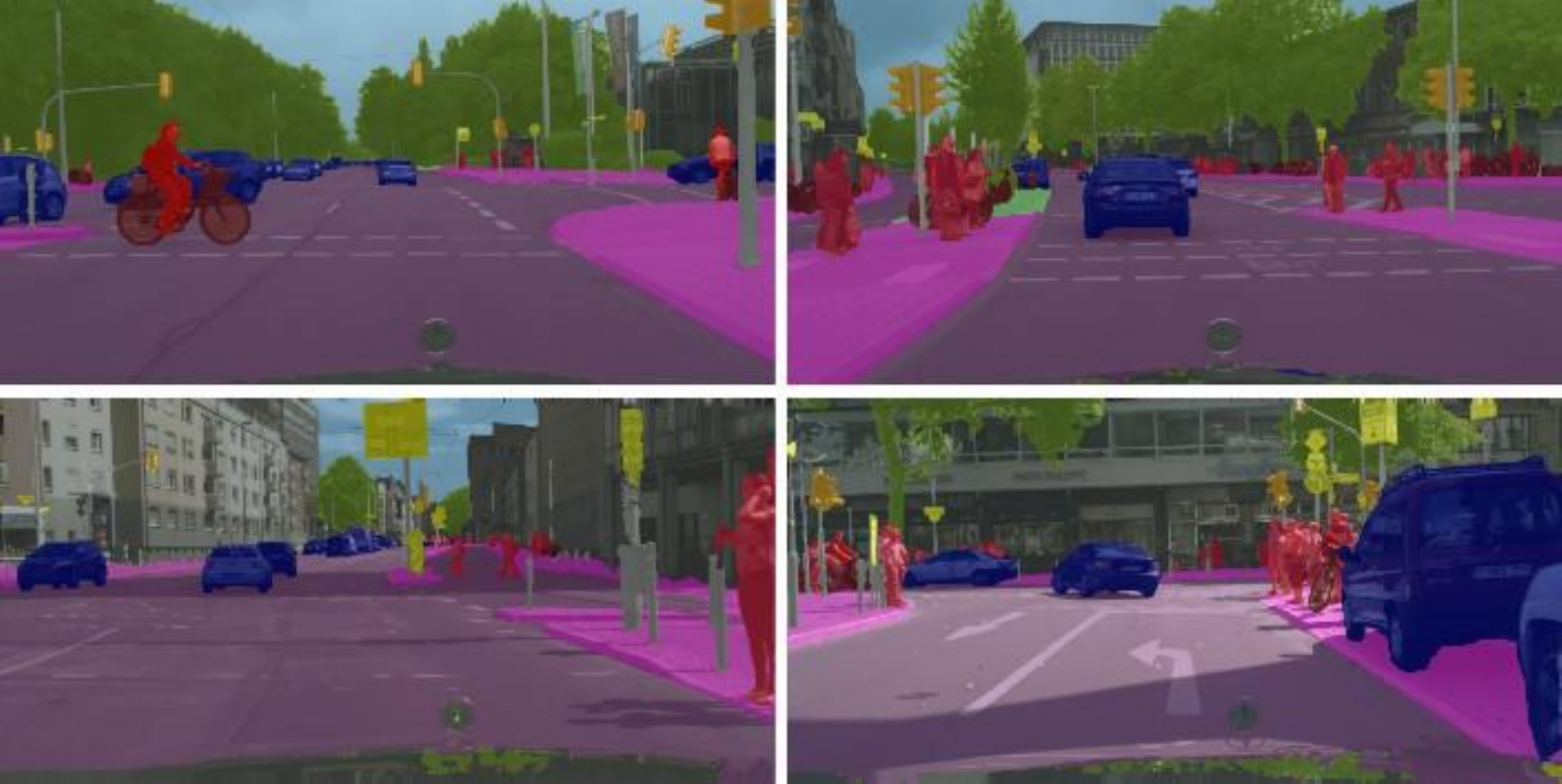}}
\caption{Part of segmentation results of SSformer on Cityscapes}
\label{fig3}
\end{figure}

\begin{table}[htb]   
\begin{center}   
\caption{Comparison of different methods on Cityscapes dataset.  }
\label{table2} 
\begin{tabular}{l l l l l}   
\hline   Method & Iteration & Params & Flops* & mIoU \\  
\hline 
FCN\cite{b8} & 80000 & 68.5M	& 619.7G & 77.02\\
DeepLabV3+\cite{b13} & 80000 & 62.6M & 571.6G & 79.88\\
DMnet\cite{b15} & 80000 & 72.2M & 627.7G & 79.19\\
PSPNet\cite{b9} & 80000 & 68.0M & 576.3G & 78.87\\
PSANet\cite{b14} & 80000 & 78.1M & 637.7G & 79.69\\
SETR-PUP\cite{b1} & 80000 & 317.3M & \textgreater362G† & 81.02\\
SSformer & 80000 & 87.5M & 355.8G & 79.84\\
\hline   
\end{tabular}   
\begin{tablenotes}
\item †Since the scale of SETR-PUP is so large that it cannot even run with the input images of 768x768 on a single RTX 3090, we cannot give precise Flops of SETR here, but it is definitely much larger than 362.1G.

\item *The Flops are calculated with the input images of 768×768 resolution except SSformer. The Flops of SSformer are calculated with the input images of 1024x1024 resolution, but the Flops is still lower than SETR.
\end{tablenotes}
\end{center}   
\end{table}

\subsection{Comparison with Other Models}
As illustrated in Table \ref{table1}, the SSformer performs better than traditional CNN-based algorithms such as FCN, PSPNet on ADE20K with much lower Flops and a little more params. Compared with latest Transformer based advanced algorithms like SETR and the original Swin Transformer with Upernet, SSformer sacrifices a little bit accuracy for much lower Flops or less Params. 

As shown in Table \ref{table2}, the results on Cityscapes are similar to
those on ADE20K. Since Cityscapes has less semantic categories and is less challenging, the performance gap between old models and latest models is not as large as that on ADE20K, but the SSformer’s advantage in terms of efficiency still exists. 

Besides, on ADE20K, SSformer is trained with a batch size 8 while the other models listed in Table \ref{table1} is trained with a batch size of 16. On Cityscapes, SSformer is trained with a batch size 4 while the other models listed in Table \ref{table2} is trained with a batch size of 8 or 16. So SSformer occupy less VRAM than other mentioned models when training. 

These quantitative results demonstrate our SSformer is a lightweight yet effective transformer for image semantic segmentation. 

\section{Conclusion and Future work}
\label{conclusion}
We presented SSformer, a lightweight transformer model for semantic segmentation. We introduce a simple MLP decoder to Transformer, which brings little complexity and parameter increase. Our SSformer takes advantage of the hierarchical multiscale architecture of Swin Transformer, and can be fine-tuned from the pretrained Swin Transformer. Experiments have proved its competitive performance and low time complexity characteristics.

We believe that its advanced features can make semantic segmentation framework more accessible to low computing power computers and embedded devices. Our source code and models are publicly available on GitHub. In the near future, we will make more experiments on different datasets and seek for further  improvement and optimization on the structure of SSformer.

\vspace{12pt}
\begin{comment}
\color{red}
IEEE conference templates contain guidance text for composing and formatting conference papers. Please ensure that all template text is removed from your conference paper prior to submission to the conference. Failure to remove the template text from your paper may result in your paper not being published.
\end{comment}

\end{document}